\documentclass[a4paper,conference]{IEEEtran}
\IEEEoverridecommandlockouts

\usepackage{cite}
\usepackage{ntheorem}
\usepackage{algorithmic}
\usepackage{graphicx}
\usepackage{textcomp}
\usepackage{xcolor}
\usepackage{hyperref}
\usepackage{mathtools}
\usepackage{dirtytalk}
\usepackage{multirow}
\usepackage{url}
\usepackage{mwe} % new package from Martin scharrer
\usepackage{caption}
\usepackage{float}
\usepackage[subpreambles=true]{standalone}
\usepackage{comment}
  \usepackage{gensymb}
\usepackage{amssymb}
\usepackage{mhchem}
\usepackage{lipsum}
\usepackage{tabularx}
\usepackage{tikz}

\usetikzlibrary{positioning}
\usepackage{float}
\usepackage{caption}%% To get \captionof

\def\BibTeX{{\rm B\kern-.05em{\sc i\kern-.025em b}\kern-.08em
    T\kern-.1667em\lower.7ex\hbox{E}\kern-.125emX}}
\begin{document}

\title{Relaxed forced choice improves performance of visual quality assessment methods%\\[5mm]
%Title tentative journal paper 2023:\\
%The relaxed alternative forced choice response format for image quality assessment*\\
\thanks{Funded by the Deutsche Forschungsgemeinschaft (DFG, German Research Foundation) -- Project ID 496858717 and DFG Project ID 251654672 -- TRR 161 (Projects A05 and C01). This research was also kindly supported by the Zukunftskolleg, the University of Konstanz, with funding from the Excellence Strategy of the German Federal and State Governments.}
%\thanks{For the purpose of a double-blind review process, information regarding funding agencies has been removed.}
}
%\begin{comment}
\author{\IEEEauthorblockN{
Mohsen Jenadeleh\IEEEauthorrefmark{1},
Johannes Zagermann\IEEEauthorrefmark{1},
Harald Reiterer\IEEEauthorrefmark{1},
Ulf-Dietrich Reips\IEEEauthorrefmark{2},
Raouf Hamzaoui\IEEEauthorrefmark{3},
Dietmar Saupe\IEEEauthorrefmark{1}}
%\author{\IEEEauthorblockN{Mohsen Jenadeleh, Johannes Zagermann, Harald Reiterer, Ulf-Dietrich Reips, Raouf Hamzaoui, Dietmar Saupe}
\IEEEauthorblockA{\IEEEauthorrefmark{1}\textit{Department of Computer and Information Science, University of Konstanz, Konstanz, Germany } }
\IEEEauthorblockA{\IEEEauthorrefmark{2}\textit{Department of Psychology, University of Konstanz, Konstanz, Germany}} 
\IEEEauthorblockA{\IEEEauthorrefmark{3}\textit{School of Engineering and Sustainable Development, De Montfort University, Leicester, UK}}
\{mohsen.jenadeleh, johannes.zagermann, harald.reiterer, reips, dietmar.saupe\}@uni-konstanz.de, rhamzaoui@dmu.ac.uk }
%{\bf \textsuperscript{*} Complete the institutions and addresses.}
%\and
%\and
%\and
%\and
%\end{comment}
%\author{\\[0.0ex]
%\IEEEauthorblockN{Anonymous QoMEX 2023 submission}\\[0.0ex]}
\IEEEoverridecommandlockouts
\IEEEpubid{\makebox[\columnwidth]{979-8-3503-1173-0/23/\$31.00
\copyright 2023 IEEE \hfill} \hspace{\columnsep}\makebox[\columnwidth]{ }}
\maketitle

\begin{abstract}
In image quality assessment, a collective visual quality score for an image or video is obtained from the individual ratings of many subjects. One commonly used format for these experiments is the two-alternative forced choice method. Two stimuli with the same content but differing visual quality are presented sequentially or side-by-side. Subjects are asked to select the one of better quality, and when uncertain, they are required to guess. The relaxed alternative forced choice format  aims to reduce the cognitive load and the noise in the responses due to the guessing by providing a third response option, namely, \say{not sure}. This work presents a large and comprehensive crowdsourcing experiment to compare these two response formats: the one with the \say{not sure} option and the one without it. 
To provide unambiguous ground truth for quality evaluation, 
subjects were shown pairs of images with differing numbers of dots and asked each time to choose the one with more dots.  
Our crowdsourcing study involved 254 participants and was conducted using a within-subject design. Each participant was asked to respond to 40 pair comparisons with and without the \say{not sure} response option and completed a questionnaire to evaluate their cognitive load for each testing condition. 
The experimental results show that the inclusion of the \say{not sure} response option in the forced choice method reduced mental load and led to models with better data fit and correspondence to ground truth. 
We also tested for the equivalence of the models and found that they were different. The dataset is available at \href{http://database.mmsp-kn.de/cogvqa-database.html}{http://database.mmsp-kn.de/cogvqa-database.html}.
%In addition, the psychometric functions fitted to the responses collected under the two test conditions are not statistically equal.
%For future image quality assessment studies applying a force choice method, we recommend to include the not-sure response option, which fulfilled our expectations by reducing mental load and leading to models with a better data fit.
%Furthermore, the resulting just noticeable difference thresholds were %larger when using the relaxed force choice format.
%we discuss the resulting just noticeable difference thresholds, which are larger when using the relaxed force choice format. 
\end{abstract} 


\begin{tikzpicture}[overlay, remember picture]

\path (current page.north) node (anchor) {};

\node [below=of anchor] {%

  2023 15th International Conference on Quality of Multimedia Experience (QoMEX)};

\end{tikzpicture}
%--------------------------------------------------------------------------------------
%--------------------------------------------------------------------------------------
%--------------------------------------------------------------------------------------
\section{Introduction}
\label{sec:introduction}
%--------------------------------------------------------------------------------------

Several testing methodologies have been established to assess the visual quality of a source image encoded with different coding parameters,  respectively bitrates. In the single stimulus presentation, the quality can be assessed by degradation category rating (DCR) on a discrete scale of five levels or with a slider on a continuous interval scale. In dual presentation mode, two images are displayed sequentially or side-by-side, and observers select the one with better quality, and when uncertain, they have to guess. From these pairwise rankings, scale values can be derived, either by scoring, as in sports rating systems, or by fitting parameters of a suitable probabilistic model~\cite{testolina2022,mantiuk2012comparison}.

The approach to let observers decide which one of two presented stimuli in a pair has a stronger perceptual effect of a certain kind is called two-alternative forced choice (2AFC). It has a long history, having been developed as a method of psychophysics by Gustav Theodor Fechner more than 150 years ago in the first edition of~\cite{fechner1889elemente}. Already back then, Fechner had relaxed the forced choice, allowing participants of his experiments to give undecided (\say{zweideutige}) responses that  he divided equally between the two alternatives before analyzing the data.

In the research area of image quality assessment, the 2AFC response format has been used almost entirely without such an option for undecided responses. In 2015, JPEG, formally known as ISO/IEC SC29 WG1, issued an international standard for procedures to test compressed images for being visually lossless when compared with the corresponding source images~\cite{ISOIEC2015}. In its Annex A and B, the forced choice paradigm was prescribed, also without a ternary choice. The first application of this standard was presented by McNally et al.~\cite{mcnally2017jpeg} for the subjective evaluation of visually lossless compressed images in the context of JPEG XS, a low-latency, lightweight video coder that is optimized for visually lossless compression. Unlike the JPEG AIC standard~\cite{ISOIEC2015}, however, an undecided response option was offered to the participants. This was justified by stating that \say{offering a ternary choice to subjects reduces subject stress and fatigue and was deemed beneficial for the reliability of the subjective evaluation results.} However, the authors did not present any experimental evidence for these claims.
%Note: This was not the *first* time, the not-sure option (akas *same*) was used for image quality assessment, as we had stated in the submission. Here are two earlier papers:
%Lee, J. S., De Simone, F., Ramzan, N., Zhao, Z., Kurutepe, E., Sikora, T., ... & Ebrahimi, T. (2010, October). Subjective evaluation of scalable video coding for content distribution. In Proceedings of the 18th ACM international conference on Multimedia (pp. 65-72).
%Lee, J. S., Goldmann, L., & Ebrahimi, T. (2013). Paired comparison-based subjective quality assessment of stereoscopic images. Multimedia tools and applications, 67, 31-48.
%Lee et al discuss several options to process the *same* responses, based on early papers in psychophysics in the 1960's or so. The Bradley-Terry model is modified and MLE used to determine parameters.

In this paper, we present the results of a large and comprehensive user study to investigate the effects of the ternary 2AFC response format for subjective studies to assess a certain visual quality of stimuli. We present paired comparisons in both response formats: 2AFC (referred to as AFC) and RFC (relaxed forced choice). In AFC, subjects are shown two visual patterns side-by-side and have to select the \say{left} or \say{right} stimulus. Subjects are instructed to guess when undecided. In RFC, participants may also select the response \say{not sure}.  
%Thus, in the following, we use the term \textit{not-sure option}. 
The focus of our study is on the following three hypotheses.

\newtheorem{hyp}{Hypothesis}
\begin{hyp}
\label{hyp_cognitive}
The not-sure option in forced choice quality assessment tasks reduces cognitive load.
\end{hyp}

The binary AFC scheme requires subjects to guess when uncertain, which may introduce more noise in the data than equally dividing the responses of the third, undecided category. It is natural to expect that noise reduction would improve the precision of the resulting estimates for scale values or parameters of fitted psychometric functions. If some form of ground truth quality is available, we can also compare the proportions of correct responses and their correlation with the ground truth qualities to assess the performance of the assessment method with and without the not-sure option.

\begin{hyp}
\label{hyp_performance}
The not-sure option in the forced choice response format improves the performance of quality assessment.
\end{hyp}

It is unclear whether the introduction of the ternary response format with the not-sure option leads to the same assessments with respect to image quality. For example, in the study for JPEG XS~\cite{mcnally2017jpeg} mentioned above, the obtained thresholds for compression parameters that yielded a just noticeable difference (JND) could have depended on the choice of the answer format of the subjective study.

\begin{hyp}
\label{hyp_homogeneity}
The psychometric functions estimated from quality assessment studies using alternative forced choice responses are the same with and without the not-sure option.
\end{hyp}

The contributions of our work can be summarized as follows:
\begin{itemize}
    \item We carried out a large crowdsourcing study with 254 participants using both the traditional two-alternative forced choice and the relaxed forced choice  formats. The data will be made available at the time of publication. 
    \item The subjectively reported \textit{Mental Demand} was significantly lower for the relaxed forced choice format. %cognitive load was partially lower for the RFC format, with a statistically significant lower \textit{Mental Demand}. %We confirmed that the relaxed forced choice format (partially) reduces the cognitive load.
    \item We show for a number of criteria that the performance of quality assessment using pair comparisons and maximum likelihood estimation of the psychometric function was improved by the relaxed forced choice format.
    \item We show that an important parameter, the JND threshold, differed significantly between the two-alternative forced choice and the relaxed forced choice formats, with a very large effect size.
\end{itemize}

%The purpose of such psychometric studies is to estimate scale values of perceptual qualities in images and video, such as technical visual quality or severity of certain artifacts. 

%--------------------------------------------------------------------------------------
%--------------------------------------------------------------------------------------
%--------------------------------------------------------------------------------------
\section{Related work}
%--------------------------------------------------------------------------------------
There have been quite a number of image and video quality assessment studies using standard two-alternative forced choice responses for paired comparison, including \cite{watson2000proposal, charrier2007maximum, miao2008quantitative, menkovski2011value, mantiuk2012comparison, ponomarenko2013new, kumcu2016performance, nuutinen2016new, zhang2018unreasonable}. On the other hand, so far, there are only very few contributions with a relaxed form of the forced choice response format \cite{mcnally2017jpeg, sun2017mdid, men2021subjective}. None of them, however, discuss or study the effect of a not-sure option in detail. However, a study by Punch et al.\ \cite{punch2001paired} in audiology compared relaxed force choice with two-alternative forced choice, revealing a higher test-retest reliability in the condition with the not-sure option.

In contrast to user studies in computer science and engineering, in psychophysics, the branch of psychology that deals with the relations between physical stimuli and mental phenomena, the unforced choice response format has been used for a very long time. Recently, the approach gave rise to new theoretical models and was tested in simulations and dedicated experiments. In 2001, Kaernbachs~\cite{kaernbach2001adaptive} proposed a Bayesian framework of a theory of indecision for experiments applying forced and unforced choice tasks. Responses to the not-sure category were split evenly. Simulations and a behavioral study with six participants were carried out. The task was to detect a brief sinusoid of 1000 Hz centered in 800 ms of white noise. By varying the signal-to-noise ratio and using an adaptive staircase procedure~\cite{kaernbach1991simple}, the absolute threshold for detection of the sinusoidal signal was estimated for each of the participants. The results showed that the unforced choice option in the adaptive procedure did not lead to reduced reliability, and a slight gain in efficiency was achieved.

Garc{\'\i}a-P{\'e}rez and Alcal{\'a}-Quintana~\cite{garcia2017indecision} presented an even more comprehensive probabilistic indecision model for dual-presentation tasks, which explicitly represents the sensory, decisional, and response components of performance by model parameters. These can be obtained from forced choice responses in pair comparisons by maximum likelihood estimation. It is claimed that the ternary response format provides more accurate estimates of model parameters than data collected with the standard binary forced choice. 
For a brief historical account of the use the ternary unforced choice, ranging from Hegelmaier and Fechner in the middle of the 19th century until it was discredited and eradicated by signal detection theory 100 years later, see~\cite{garcia2019Thedos}.

The intense research in psychophysics on the implications of allowing --- or rather ruling out --- a not-sure option in forced choice experiments is encouraging for applications in multimedia quality assessment. 
While in psychophysics the focus lies primarily on the individual performance of a subject, it is just the opposite in multimedia quality assessment; the collective experience of quality has to be assessed. 
Therefore, it is unclear whether the conclusions drawn in the recent works in psychophysics can simply be transferred one-to-one to our domain of research. 

\section{Experimental design and setup}
%--------------------------------------------------------------------------------------
\begin{comment}
In our experiment, we used dot images with a relatively large number of dots to prevent participants from counting the dots. These images are suitable for our experiment because:
\begin{itemize}
\item The correct ranking order of the dot images is known, allowing us to objectively assess the subjects' responses.
 
 \item  The task of comparing two dot images to determine which one has more dots is simple to understand, reducing the chance of unreliable responses due to misunderstandings of the subjective task.

 \item The number of dots between paired images varies, with some pairs being harder to compare than others. For instance, the number of dots can differ by only two in some pairs, while in others it can vary by 40. This introduces uncertainty in the assessment of some paired images, making it interesting to study the impact of offering a \say{not sure} option in addition to selecting the left or right image.
\end{itemize}
\end{comment}

For our experiments, we have chosen the so-called dot-guessing game \cite{horton2010dot}. In this game, subjects are presented with pairs of images consisting of black dots of the same size, but with differing numbers, for example, 300 and 320. The question asked is which of the two images in a pair contains more dots. Thus, the number of dots is regarded as the \say{quality} of an image that is to be assessed. %For example, in our experiment, one of the two images always had 300 dots, and we found the difference threshold or JND to be about 26 dots.

Why are we not using source images compressed at different bitrates? It has been observed that slightly compressed images may be judged to have better visual quality than the source images. This may be due to the denoising effect of some compression methods at high bitrates. This is also the reason why in the paired comparisons in Annex A of the JPEG standard~\cite{ISOIEC2015}, the reference image is displayed a second time above the pair so that the question for better image quality could be recast as \say{Please select the lower image that is the closest match to the reference.} 

With the dot images, we do not have this problem and may safely assume that the perceived number of dots on the observer's sensory scale is monotonically related to the number of dots in the stimulus. In expectation, more dots will be perceived when the number of dots is increased. This is an advantage that offers a wider spectrum of criteria when comparing assessment performance between the AFC and RFC response formats.

 Moreover, random dot images have been used in subjective studies for various other applications, including human computation tasks~\cite{horton2010dot}, aggregate predictions using the wisdom of crowds~\cite{honda2022round,ugander2015wisdom,kemmer2020enhancing}, and information aggregation~\cite{pfeiffer2012adaptive}. So these works have shown that the dot guessing game may indeed be considered as a basic task that can serve as a \say{fruit fly for human computation research} as Horton has expressed it \cite{horton2010dot}.

We conducted a crowdsourcing user study to collect responses for the AFC and RFC paired comparisons, using a within-subjects design in which all participants responded to both conditions. The presentation order of AFC and RFC was randomized to avoid the potential for presentation order bias. 

We measured participants' cognitive load using the NASA task load index (TLX) questionnaire~\cite{hart1988development}. The TLX is a multi-scale questionnaire to assess participants' perceived subjective workload -- split into six sub-scales:  \textit{Mental Demand}, \textit{Physical Demand}, \textit{Temporal Demand}, \textit{Performance}, \textit{Effort}, and 
 \textit{Frustration}. Although the original analysis procedure includes an individual weighting of the sub-scales, it is common to skip the weighting and only average participants' ratings for the sub-scales to create scores for each sub-scale and the overall task load (known as \textit{raw} TLX)~\cite{10.1145/3582272}. Scores rank between 0 and 100; lower values indicate a lower task load.

\subsection{Image generation and procedure}

In our experiment, each dot image contained a relatively large number of non-overlapping dots and had a resolution of $640\times480$ pixels to facilitate side-by-side display on crowdworkers' screens. We defined a \textit{question} as showing a participant a pair of images side-by-side and asking to select the image that contains more dots. We compared a reference image with 300 dots to 20 test images with a greater number of dots, ranging from 302 to 340 with a step size of 2. For each question, the reference image was presented on either the left or the right side, with the test image on the opposite side. This resulted in 40 study questions for each of the two conditions, AFC and RFC, for a total of 80 study questions presented to each subject.  Each image was generated with a random dot pattern.

To mitigate the influence of learning from previous comparisons or random dot patterns on participants' responses, we ensured that no dot pattern was repeated in the 80 study questions evaluated by a subject in the two test conditions. For example, the 80 reference images, each containing 300 dots, had unique patterns. Moreover, the dot patterns also differed between subjects. The order of the conditions used, AFC or RFC, was also randomly assigned to participants. 

\begin{figure}[t!]
\centering
\begin{minipage}{0.98\linewidth}
\resizebox{0.99\textwidth}{!}{\includegraphics[width=0.98\linewidth]{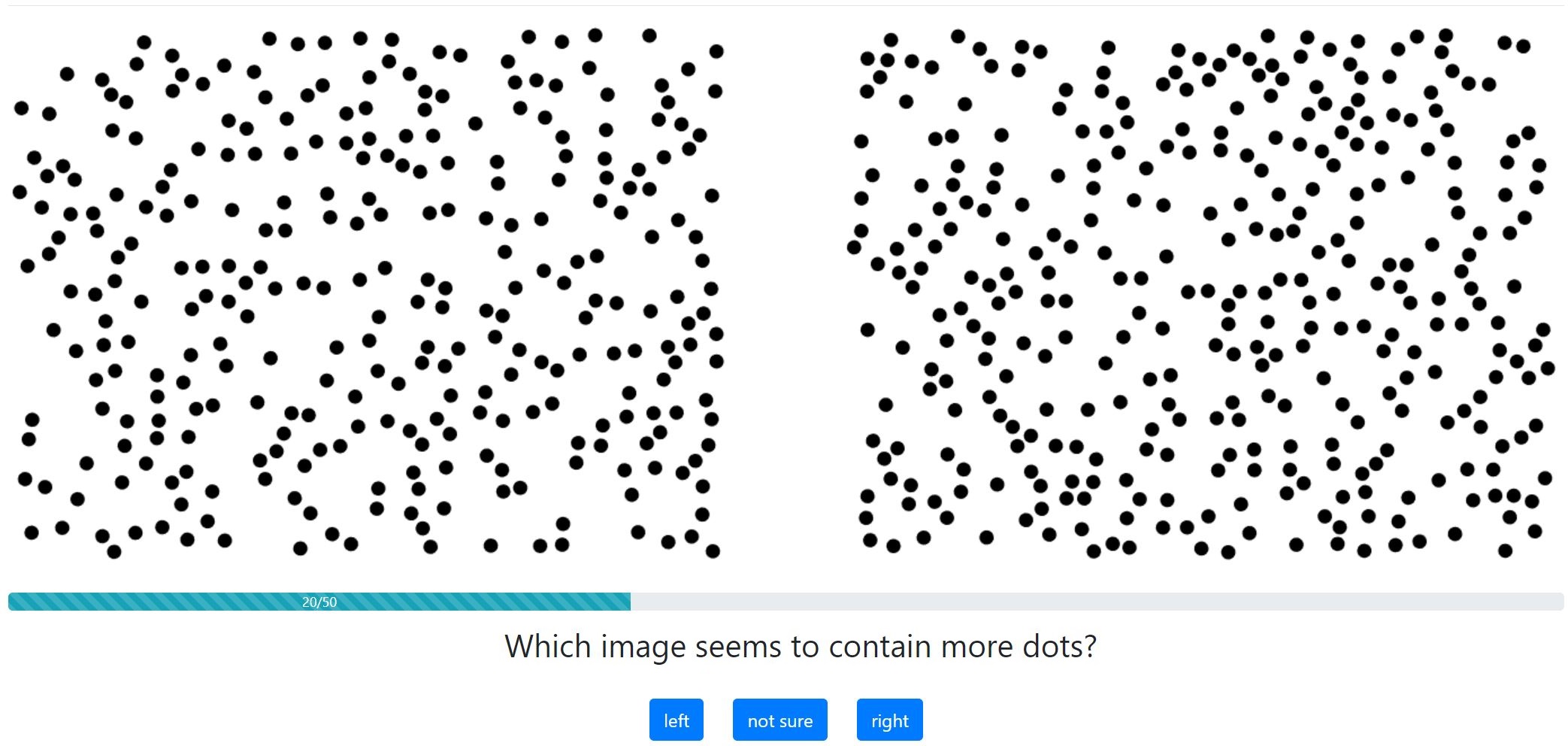}}
\end{minipage}
\caption{User interface with relaxed forced choice answer options and a progress bar.}
\label{fig:interface}
\end{figure}

%For each test image with a greater number of dots, the participant compared four images with the same number of dots to the reference image in both test conditions, each with a unique random pattern. 
%To further reduce the potential impact of random dot image patterns on participants' ratings, 80 sets of images were generated, each consisting of 80 study images as described. The pattern of each image in these 80 sets was unique and did not repeat in other sets. 
 % As a result, approximately half of the participants did the AFC experiment first, followed by the RFC experiment, while the other half did the reverse.

To identify and filter out unreliable participants, e.g., line clickers, we included 10 trap questions for each test condition, comparing a reference image with 300 dots to a test image with about 450 dots. In these trap questions, the test image was easily distinguishable. Each participant was presented with 10 trap questions in each experimental condition. 
 
\begin{table*}[t!]
\centering
\caption{Summary of the responses to the study questions in the two test conditions collected from 235 participants.} 
\resizebox{1\textwidth}{!}{
\large
 \begin{tabular}{*{21}{c} c@{\hspace{0.5cm}}}
Test condition & Number of dots & 302 & 304 & 306 & 308 & 310 & 312 & 314 & 316 & 318 & 320 & 322 & 324 & 326 & 328 & 330 & 332 & 334 & 336 & 338 & 340 \\ \hline
\multirow{2}{*}{AFC}& Correct answers & 254 &  256  & 284 &  269  & 306  & 294 &  309 &  320 &  318 &  339 &  359 &  370 &  349 &  367 &  383 &  374 &  382 &  366  & 392 &  404\\ 
& Wrong answers  & 210 &  208 &  183 &  197 &  160 &  175 &  158&   149 &  150  & 129  & 108 &   97 &  118   & 99  &  83  &  93  &  85 &  101 &   70  &  64
\\ 
\hline 
\multirow{3}{*}{RFC} & Correct answers &210  & 230  & 239 &  251 &  254  & 264  & 274  & 285  & 306  & 303 &  323 &  319 &  321 &  330   &362  & 350  & 352  & 385 &  357  & 377
\\  
& \say{not sure} answers   &  64   & 53   & 70  &  62  &  56  &  54  &  61  &  61  &  49  &  45 &  39  &  45 &   40  &  38  & 39  &  31 &   35  &  22  &  38  &  29 \\ 
& Wrong answers &190  & 181  & 158 &  153 &  156 &  151 & 132 &  123 &  113&   120 &  105  & 103 & 106 &   98 &   65  &  86  &  80  &  60  &  67 &   62\\ \hline 
\end{tabular}
}
\label{Table:data}
\end{table*}
\begin{comment}
    
\begin{table*}[t!]
\centering
\caption{Summary of the responses to the study questions in the two test conditions collected from 235 participants.} 
 \resizebox{1\textwidth}{!}{
\begin{tabular}{*{21}{c@{\ }}c}
Test condition & Number of dots & 302 & 304 & 306 & 308 & 310 & 312 & 314 & 316 & 318 & 320 & 322 & 324 & 326 & 328 & 330 & 332 & 334 & 336 & 338 & 340 \\ \hline
\multirow{2}{*}{AFC}& Correct answers & 254 &  256  & 284 &  269  & 306  & 294 &  309 &  320 &  318 &  339 &  359 &  370 &  349 &  367 &  383 &  374 &  382 &  366  & 392 &  404\\ 
& Wrong answers  & 210 &  208 &  183 &  197 &  160 &  175 &  158&   149 &  150  & 129  & 108 &   97 &  118   & 99  &  83  &  93  &  85 &  101 &   70  &  64
\\ 
\hline 
\multirow{3}{*}{RFC} & Correct answers &210  & 230  & 239 &  251 &  254  & 264  & 274  & 285  & 306  & 303 &  323 &  319 &  321 &  330   &362  & 350  & 352  & 385 &  357  & 377
\\  
& \say{not sure} answers   &  64   & 53   & 70  &  62  &  56  &  54  &  61  &  61  &  49  &  45 &  39  &  45 &   40  &  38  & 39  &  31 &   35  &  22  &  38  &  29 \\ 
& Wrong answers &190  & 181  & 158 &  153 &  156 &  151 & 132 &  123 &  113&   120 &  105  & 103 & 106 &   98 &   65  &  86  &  80  &  60  &  67 &   62\\ \hline 
\end{tabular}
}
\label{Table:data}
\end{table*}
\end{comment}

\subsection{Crowdsourcing study}
%In our crowdsourcing experiment, we collected paired comparison responses using the two-alternative forced-choice (AFC) method and a method with a relaxed \say{not sure} response option in addition to the \say{left} and \say{right} image choices (RFC) to investigate the research questions listed in section \ref{sec:introduction}. 

%We conducted our experiment on the Amazon Mechanical Turk (MTurk) platform. We posted a HIT (human intelligence task) with 254 assignments. Each MTurk worker was allowed to perform only one assignment.
We conducted our experiment on the Amazon Mechanical Turk (MTurk) platform. We posted a human intelligence task (HIT) with 254 assignments. However, each MTurk worker was only allowed to do one assignment. In each assignment, a worker had to answer 50 questions (40 questions for the study experiment and 10 trap questions) for the AFC experiment and 50 questions for the RFC experiment. The order of the experiment types and the order of the questions in each experiment were randomized. Figure~\ref{fig:interface} shows a screenshot of the user interface.

To ensure reliable responses, participants were required to have at least 500 previously approved HITs with a 95\% or higher approval rate on MTurk. Additionally, participants were required to use a PC or laptop with a minimum logical resolution of $1366 \times 768$ pixels and the Google Chrome browser to complete the experiment. Participants who did not meet these requirements received a warning message, and the experiment was terminated.

A brief instruction was shown to the eligible workers, explaining the steps of the experiment, and the worker was asked to sign a consent form. The presentation order of the two conditions, AFC and RFC, was chosen randomly. More detailed instructions, the training session, and the main part of the experiment were given for the first chosen condition, and then this was repeated for the other condition.

For each test condition, participants received detailed instructions with examples of paired comparisons specific to that test condition, followed by an explanation of the NASA task load questionnaire.  %These instructions included a challenging example of a pairwise comparison where there was a small difference in the number of dots between the reference image and the test image. They also included an example of a pairwise comparison where there was a relatively large difference in the number of dots between the two compared images. 
%Then, the NASA Task Load Questionnaire was explained to the worker. %Next, the worker was asked to answer the questionnaire after doing the experiment, taking into  consideration only the pairwise comparison questions in the current test condition. 
After reading the instructions, workers went through a training session consisting of five questions of varying levels of difficulty. %, where the task was to compare pairs of dot images  
%to familiarize the worker with the experiment.
After answering each training question, the worker was given feedback indicating which image had more dots. 

Then the worker was asked to do the main experiment with 50 questions for the first selected test condition (40 study and 10 trap questions).  For each question, the paired images were shown for five seconds. If the worker did not make a decision during these five seconds, a blank page was displayed in place of the paired images, and the worker was given three more seconds to answer.  If the worker did not give an answer, the next question was shown. After the 50 questions, the NASA task load questionnaire was shown. After answering the questionnaire, the experiment for the other condition was carried out in the same manner. %Fig.~\ref{fig:exp_steps} shows the steps of the experiment. 

The experimental procedures and protocols used in the study were ethically approved by the Institutional Review Board of the local university.%  University of .... (sentence will be revised).
 
%\begin{figure}[!t]
%\centering
%\begin{minipage}{0.98\linewidth}
%\resizebox{0.99\textwidth}{!}{\includegraphics[width=0.98\linewidth]{notsure_figures/steps3.jpg}}
%\end{minipage}
%\caption{Steps of the subjective study when the AFC experiment is %randomly selected to be conducted first.}
%\label{fig:exp_steps}
%\end{figure}

\section{Experimental results}
In the crowdsourcing study, 254 crowdworkers responded to 25,400 paired comparisons and provided 3,048 answers to the NASA TLX questions. The response times for the paired comparisons were recorded. In this section, we present the filtering of unreliable participants, the data analysis for our hypotheses.%, and the response times.

%old version
%We recruited 254 workers by posting a HIT with 254 assignments on Amazon Mechanical Turk. For each of the two test conditions, each worker answered 40 study questions, 10 trap questions, and completed the NASA TLX questionnaire with six questions. In total, we collected 25,400 paired comparisons and 3,048 answers to the NASA TLX questions. The response times for the paired comparisons were also recorded. 

\subsection{ Data and model}

\subsubsection{Data cleansing}
We used two criteria to eliminate unreliable participants. First, the responses of participants who answered incorrectly three or more trap questions were discarded. Second, the responses of participants who skipped five or more questions from the 100 (study and trap) questions in the main study were disregarded. Consequently, 19 participants were excluded, and their responses were not considered in the analysis. The analysis was conducted based on the responses of the remaining 235 participants.
%We retained responses from participants who may have skipped up to four questions. 
In order to maintain a strict within-subject design, if a participant did not answer a study question in one test condition, the corresponding question from the other test condition was excluded. The summary of the remaining data is presented in Table~\ref{Table:data}.

\subsubsection{Model and MLE fitting}
\label{model_fit}
\begin{figure}[tp!]
\centering
%\begin{minipage}{0.49\linewidth}
 %\resizebox{1\textwidth}{!}{\includegraphics[width=0.98\linewidth]{notsure_figures/fitting_AFC_235_normal_CI_bar.png}}\\
%\centering
%(a) AFC
%\end{minipage}
%\begin{minipage}{0.49\linewidth}
 %\resizebox{1\textwidth}{!}{\includegraphics[width=0.98\linewidth]{notsure_figures/fitting_RFC_235_normal_CI_bar.png}}\\
%\centering
%(b) RFC
%\end{minipage} 
\begin{minipage}{0.95\linewidth}
\resizebox{1\textwidth}{!}{\includegraphics[width=0.98\linewidth]{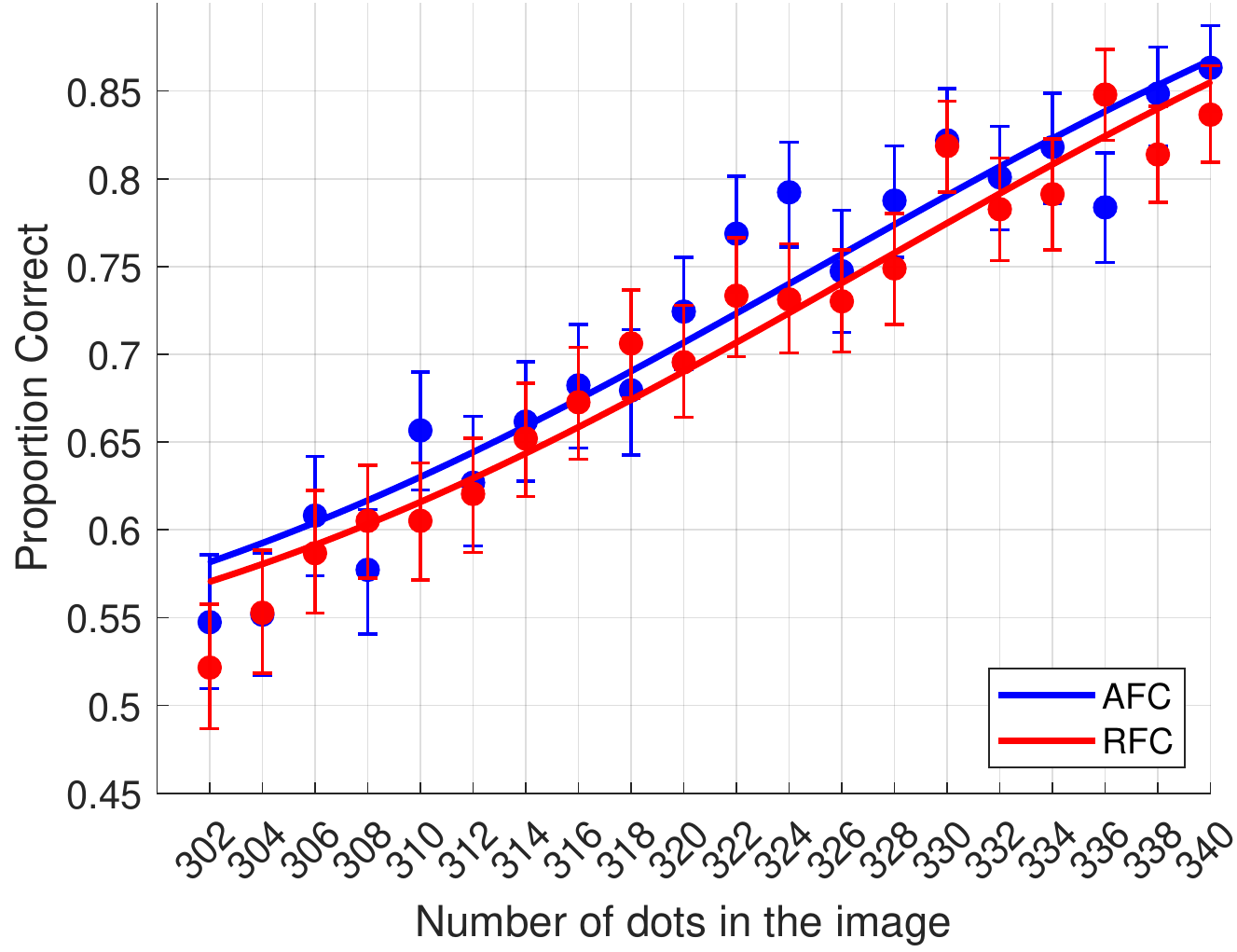}}\\
\centering
\end{minipage} 
\vspace{-10pt}
\caption{The psychometric function in Equation (\ref{eqn:PSF}) is fitted to the proportions of correct responses in each test condition.}
\label{fig:fitting}
\end{figure}

In psychophysics, maximum likelihood estimation (MLE) is applied to fit a psychometric function to the proportions of correct paired comparison responses. Commonly, the cumulative distribution function $\Phi(x; \mu, \sigma)$ of a normal distribution is used as follows \cite{prins2016Psychophysik},
\begin{equation}
\label{eqn:PSF}
 \psi(x;\mu, \sigma) = \frac{1}{2} + \frac{1}{2} \Phi(x; \mu, \sigma).
\end{equation}
The additive constant 1/2 is required here to accommodate the guessing, which can be expected to give the correct answer half of the time.
To analyze the responses collected through the RFC method, we divided the \say{not sure} responses into two halves, correct and incorrect, as usual. Figure~\ref{fig:fitting} shows the resulting psychometric functions.
The parameters for the psychometric functions were $\mu = 25.18, \sigma = 23.61$ for AFC, and $\mu = 27.10, \sigma = 23.33$ for RFC. Thus, the estimated JNDs were 25.18 and 27.10, respectively.

To determine the confidence intervals (CI) of the proportions of correct responses, non-parametric bootstrapping with 1000 trials was used. The 235 subjects were sampled with replacement. For each bootstrap sample, all of the responses of the sampled subjects were included in computing the proportions of correct responses, thereby preserving the within-subject design. The percentile bootstrap intervals for 1000 trials produced 95\% CIs for the 20 proportions (Figure \ref{fig:fitting}).

%To compare the equality of the fitted psychometric functions based on the responses, we utilized the non-parametric Split Mantel-Haenszel test proposed by Garc{\'\i}a et al. \cite{garcia2018nonparametric}.
 
\begin{comment}

\begin{table*}[t!]
\centering
\caption{McNemar test results for paired proportions at each stimulus level in the two test conditions.} 
\resizebox{\textwidth}{!}{%\resizebox{18 cm}{!}{
\Huge
\begin{tabular}{*{20}{c@{\ }}c}
%\begin{tabular}{cccccccccccccccccccccc}
Number of dots & 302 & 304 & 306 & 308 & 310 & 312 & 314 & 316 & 318 & 320 & 322 & 324 & 326 & 328 & 330 & 332 & 334 & 336 & 338 & 340 \\ \hline
Chi square & 0.504 &  0.001 &    0.362  &   0.679   &   2.784   & 0.019  &  0.061  &  0.059   &  0.687  &  0.856  &    1.516 &    5.025 &    0.269  &    1.806  & 0.002  &  0.418   &  1.054   &   6.276  &     1.875   &   1.155\\ 
p-values & 0.478  &   0.972  &   0.548  &   0.410 &   0.095  &   0.89    & 0.806  &   0.808  &   0.407 &   0.354  &  0.218  &  0.025  &   0.604 &   0.179 &    0.965  &  0.518   &  0.305 &   0.012   &   0.171   &   0.282\\ 
\end{tabular}
}
\label{Table:McNemar}
\end{table*}
    
\end{comment}
 \begin{table*}[t!]
\centering
\caption{McNemar test results for paired proportions at each stimulus level in the two test conditions.} 
 \resizebox{1\textwidth}{!}{
%\begin{tabular}{*{20}{c@{\ }}c}
\begin{tabular}{*{20}{c@{\hspace{0.2cm}}}c}
Number of dots & 302 & 304 & 306 & 308 & 310 & 312 & 314 & 316 & 318 & 320 & 322 & 324 & 326 & 328 & 330 & 332 & 334 & 336 & 338 & 340 \\ \hline
\multirow{1}{*}{} Chi square & 0.504 &  0.001 &    0.362  &   0.679   &   2.784   & 0.019  &  0.061  &  0.059   &  0.687  &  0.856  &    1.516 &    5.025 &    0.269  &    1.806  & 0.002  &  0.418   &  1.054   &   6.276  &     1.875   &   1.155\\ 
 \multirow{1}{*}{} p-values & 0.478  &   0.972  &   0.548  &   0.410 &   0.095  &   0.89    & 0.806  &   0.808  &   0.407 &   0.354  &  0.218  &  0.025  &   0.604 &   0.179 &    0.965  &  0.518   &  0.305 &   0.012   &   0.171   &   0.282\\ 
\end{tabular}
}
\label{Table:McNemar}
\end{table*}

 \begin{figure*}[!ht]
\centering
\begin{minipage}{0.245\linewidth}
%\resizebox{1\textwidth}{!}{\input{figures/ }}\\
\resizebox{1\textwidth}{!}{\includegraphics[width=0.98\linewidth]{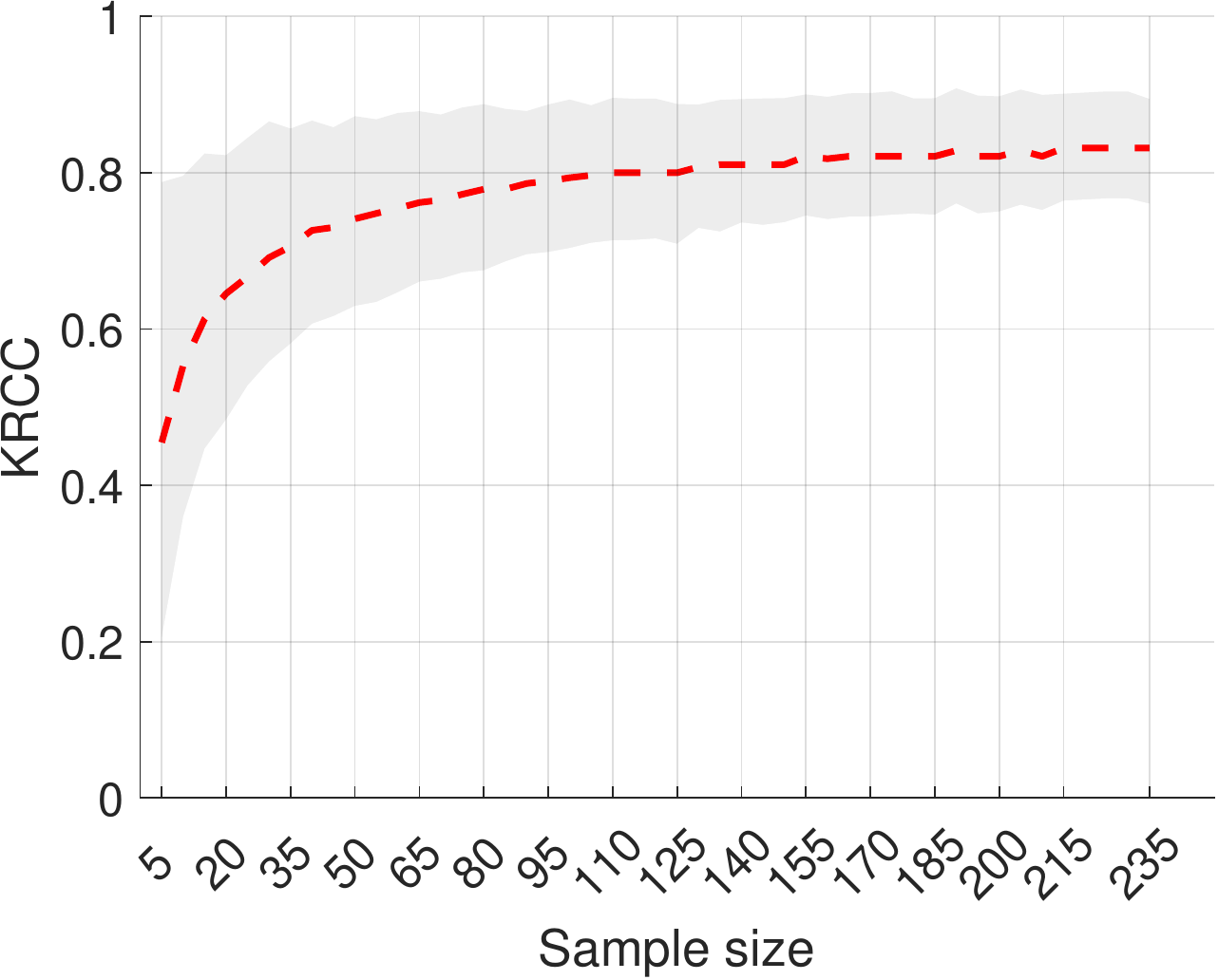}}\\
\centering
(a) AFC
\end{minipage}
\begin{minipage}{0.245\linewidth}
%\resizebox{1\textwidth}{!}{\input{figures/ }}\\
\resizebox{1\textwidth}{!}{\includegraphics[width=0.98\linewidth]{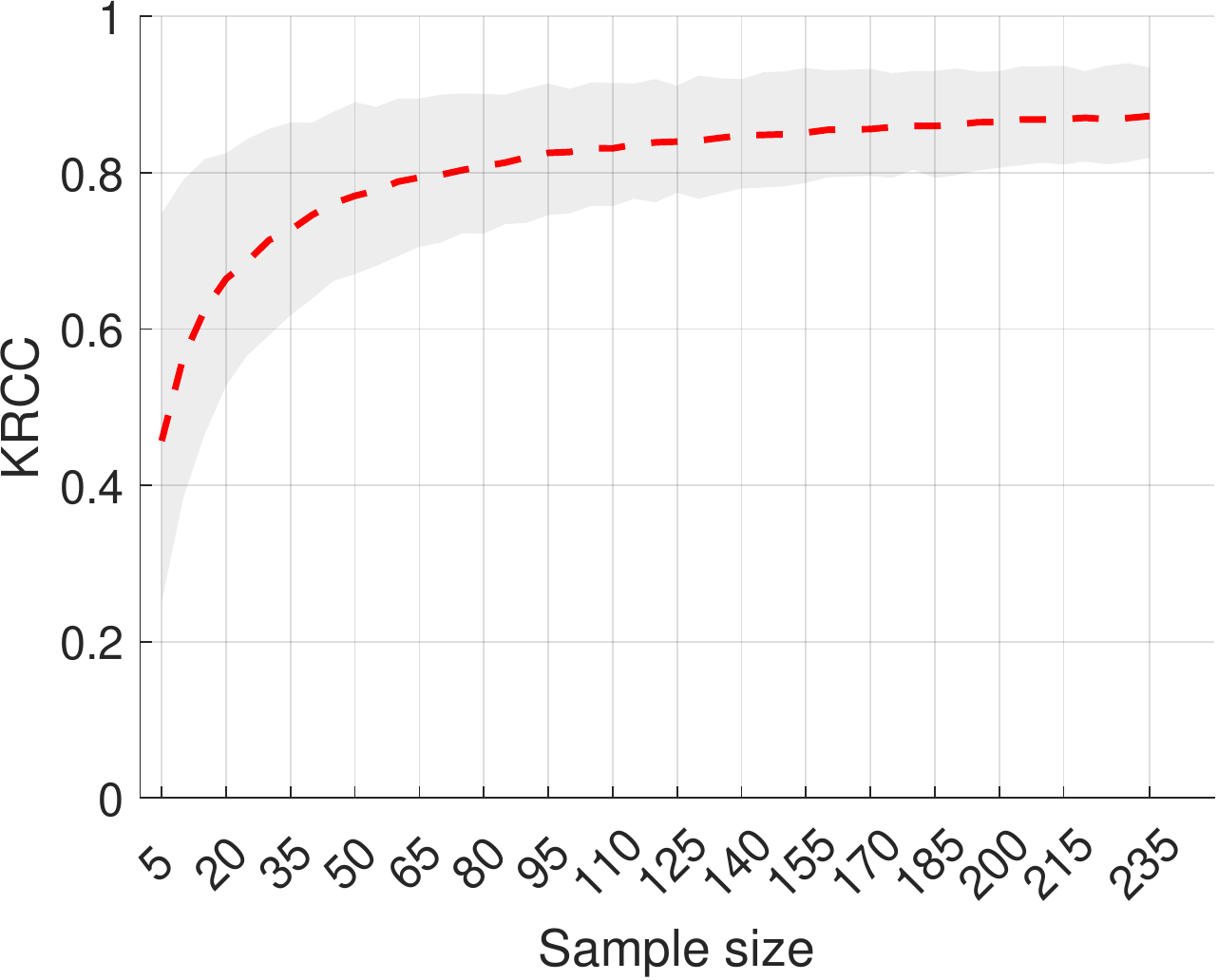}}\\
\centering
(b) RFC
\end{minipage}
\begin{minipage}{0.245\linewidth}
%\resizebox{1\textwidth}{!}{\input{figures/ .tifD}}\\
\resizebox{1\textwidth}{!}{\includegraphics[width=0.98\linewidth]{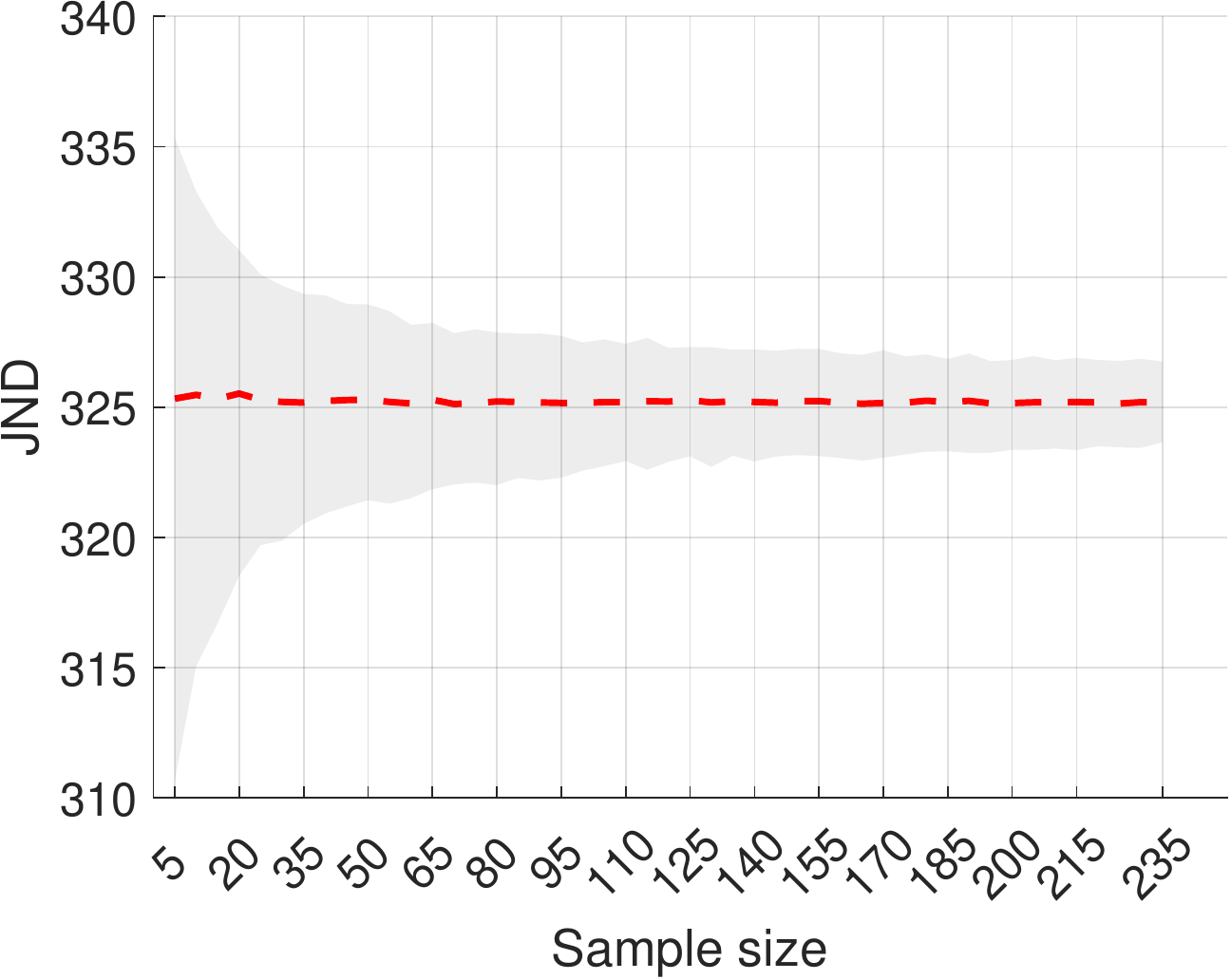}}\\
\centering
(c) AFC
\end{minipage}
\begin{minipage}{0.245\linewidth}
%\resizebox{1\textwidth}{!}{\input{figures/ .tif}}\\
\resizebox{1\textwidth}{!}{\includegraphics[width=0.98\linewidth]{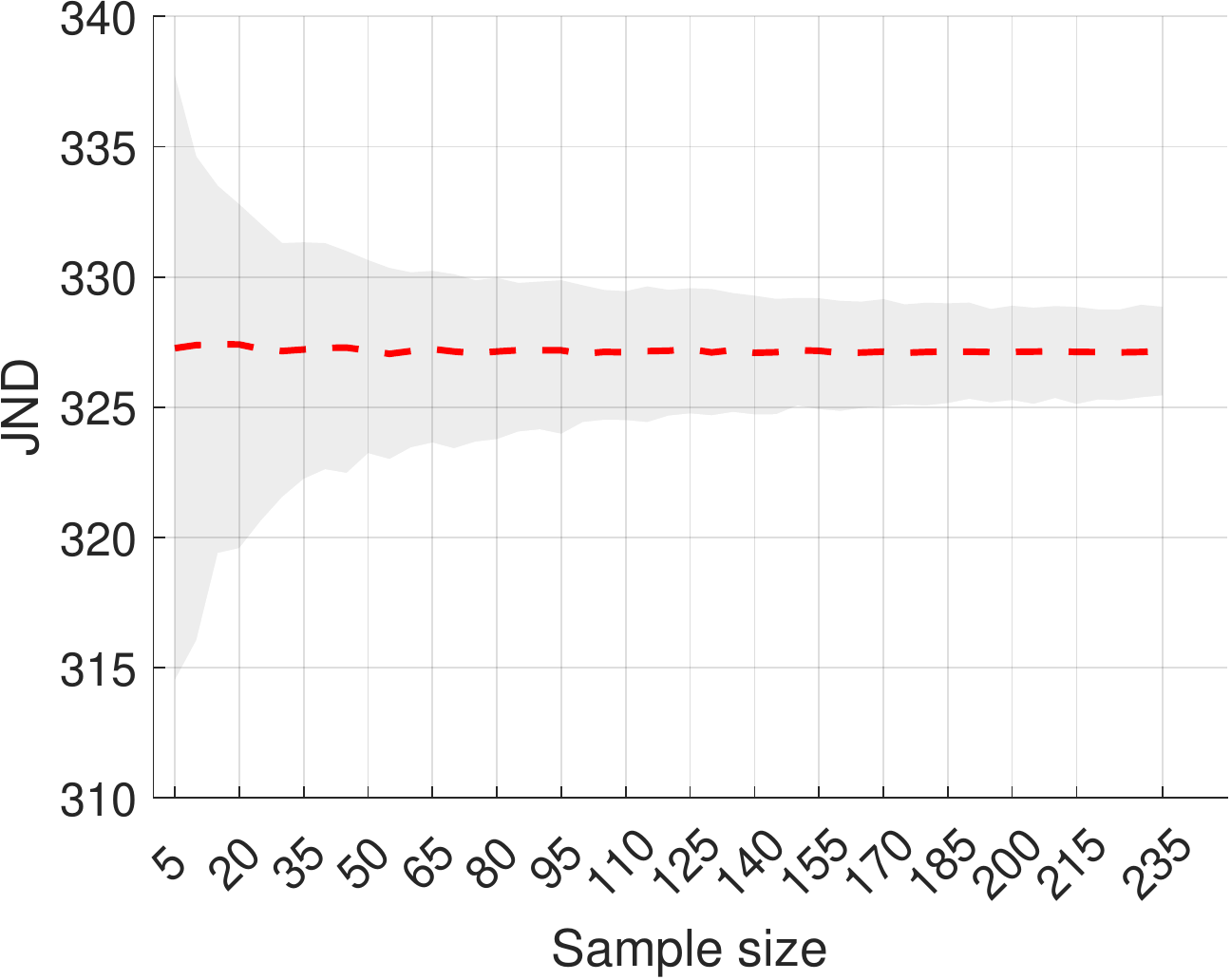}}\\
\centering
(d) RFC
\end{minipage}
\caption{The first two plots illustrate the correlation (KRCC) between the proportions and the stimulus levels, while the second two plots illustrate the estimated JND values, both with 95\% CI. These statistics were calculated using 1000 bootstrap samples.}
\label{fig:KRCC_JND}
\end{figure*} 

\subsection{Cognitive load analysis (Hypothesis~\ref{hyp_cognitive})}
We used a Wilcoxon matched pairs signed ranks test to analyze the overall TLX scores and the sub-scales. The analysis of the overall TLX scores did not show statistically significant differences (z = $-$1.65, p \textgreater~.05) with median values of 57.50 for AFC and 57.50 for RFC. The analysis of the sub-scale \textit{Mental Demand} revealed a statistically significant difference between AFC and RFC (z = $-$2.19, p \textless~.05), showing higher median values regarding Mental Demand for AFC (77.50) than for RFC (75.00). None of the other sub-scales showed a statistically significant difference between the conditions.
%with mean values of 56.35 for AFC and 55.48 for RFC. The analysis of the sub-scale \textit{Mental Demand} revealed a statistically significant difference between AFC and RFC (z = $-$2.19, p \textless~.05), showing higher mean values regarding Mental Demand for AFC (70.61) than for RFC (68.03). None of the other sub-scales showed statistically significant differences between the conditions.

\subsection{Performance analysis (Hypothesis~\ref{hyp_performance})}
We checked whether the not-sure option in our forced choice experiment with the dot-guessing game improved the goodness of fit of the psychometric function, increased the correlation with the ground truth ordering of the stimuli according to dot numbers, and reduced the confidence intervals for the estimated JND. To indicate the degree by which some of these criteria differed between AFC and RFC, we used Cohen's effect size $d$ for paired observations~\cite{lakens2013calculating}.

\subsubsection{Goodness of fit}
The goodness of fit assesses the accuracy of the fitted psychometric function in predicting the proportion of responses in each condition. To evaluate the goodness of fit of the psychometric function, we applied the procedure of analysis by bootstrapping given by Wichmann and Hill \cite{wichmann2001psychometric}. The deviance was calculated from the observed and predicted proportions at each of the 20 stimuli levels for each condition and summed to form the overall metric values. The total deviance value was 37.00 for the AFC condition and 23.31 for the RFC condition. 

A smaller deviance indicates  better goodness of fit. From an analysis with 1000 bootstrap samples, we also calculated the p-values and found that they were equal to 0.008 for AFC and 0.270 for RFC. The p-value is the estimated probability that a random sample of the psychometric model leads to a deviance greater than that from the observed data. Thus, the p-value of only 0.008 for AFC suggests that the AFC model should be rejected. Furthermore, the difference in deviance between AFC and RFC was large, with an effect size of 0.98.

%Effect size interpretations accorimng to Cohen and Sawilowsky (fram Wikipedia)
%Very small	0.01
%Small	0.20
%Medium	0.50
%Large	0.80
%Very large	1.20
%Huge	2.0

%The Pearson $\chi^2$ values were 37.24 for the AFC data and 23.11 for the RFC. 

%To measure the improvement of the RFC model compared to the AFC model through the incorporation of the \say{not sure} response in term of deviance metric, we calculated the $R^2 = 1- \frac{D_{RFC}}{D_{AFC}}$ statistic, where $D_{AFC}$ and $D_{RFC}$ are the overall deviance for each test condition. The calculated $R^2$ value was 0.33.
%Our analysis shows that the RFC experiment provided a better goodness of fit, as indicated by the lower deviance value. These results suggest that a more accurate assessment of the psychometric function's goodness of fit can be achieved with RFC.

\subsubsection{Correlation with ground truth}
As explained in Section~\ref{sec:introduction}, we can use the rank order correlation between the observed proportions of correct responses and the numbers of dots in the test stimuli to assess the performance of AFC and RFC. Methods with larger correlations can be considered more efficient. We have used Kendall's rank order correlation (KRCC) which is less sensitive to ties and assumptions about the distribution than  Spearman's rank order correlation.

%Kendall's rank order correlation coefficient (KRCC) is a non-parametric method for measuring ordinal association. It calculates the number of concordant and discordant pairs in a set of data and is less sensitive to ties and assumptions about the distribution. Hence, we utilized it to report the correlation between proportions and their physical scale. With this more accurate measurement method, we anticipate that as the stimuli level (number of dots) increases, the proportion of participants detecting the test images will increase. On the other hand, with a small difference between the reference and test stimuli, the choices will be more random, or \say{not sure} will be selected if available.

Fig. \ref{fig:KRCC_JND} (a, b) shows the mean KRCC values with 95\% CIs, computed for sample sizes from 5 to 235 from 1000 bootstrap samples as outlined in Section~\ref{model_fit}. The correlation for RFC was larger and had smaller CIs. For the maximal sample size of 235, the mean KRCC value of 1000 bootstrap samples for AFC was 0.83 with a CI length of 0.13, while the KRCC for RFC was 0.87 with a CI length of 0.11. The difference in correlation between AFC and RFC was of medium effect size, 0.79. Thus, in terms of correlation with ground truth, the quality assessment with the not-sure option was superior to the alternative forced choice.

\subsubsection{Convergence and precision of the JND}
Fig.~\ref{fig:KRCC_JND} (c,d) shows the JND estimates and their CIs for various sample sizes computed using 1000 bootstrap samples again. For the sample size 235, the CI of the JND was 3.11 for AFC and 3.40 for RFC. The AFC experiment resulted in a slightly smaller CI length.%, but this difference may not have significant practical implications. Visual inspection of the CIs across different sample sizes show no substantial variations.

\subsection{Homogeneity of psychometric functions (Hypothesis~\ref{hyp_homogeneity})}
We applied three criteria for testing the homogeneity of the models derived from AFC and RFC data, one for each of the predicted proportions of correct responses, one for all of them together, and one for equality of the estimated JNDs. 

\noindent (1) For the first task, we conducted the McNemar test for the statistical analysis of the 2x2 contingency tables that summarize the binary responses for each stimulus level. % The 2x2 matrix is used to evaluate if there is a significant difference in the proportion of \say{same} versus \say{different} responses between two conditions. 
The test statistic is used to test the null hypothesis of no difference.
The resulting chi-square and p-values are presented in Table~\ref{Table:McNemar}. The critical value at a 95\% significance level was 3.84. If the McNemar chi-square value is greater than the critical value, it indicates a significant difference with 95\% confidence. 
The results suggest that there was weak evidence for rejecting the null hypothesis of equality for 18 paired proportions in the two test conditions. However, two paired proportions showed a significant difference.

\noindent (2) To compare the psychometric functions fitted to the proportions of correct responses at all stimuli levels together, we applied the nonparametric generalized Mantel-Haenszel test \cite{garcia2018nonparametric}. The p-value was 0.02, rejecting the null hypothesis of homogeneity of the psychometric functions in the compared test conditions.

\noindent (3) Finally, the difference between the JNDs estimated with AFC and RFC was significant as the effect size was huge, estimated at 2.29 from 1000 bootstrap samples. 

In summary, the statistical tests conducted indicate that the data collected with the AFC and RFC were significantly different, and Hypothesis~\ref{hyp_homogeneity} has to be rejected.

%\subsection{Response time analysis}
%Response time was recorded as the duration between the time the stimuli were displayed and the time the button was pressed (either \say{left}, \say{right}, or \say{not sure}). The mean response time for the questions in the RFC experiment was 2.19 s, while the mean response time in the AFC experiment was 2.08 s. However, this difference might not be important in practice. 

\section{Conclusions and future work}
This study examined the effects of including the \say{not-sure} response option for pair comparisons on subjects' cognitive load and efficiency. The task was to assess the perceptual quality for a set of visual stimuli of dot patterns that may well serve as a paradigm for visual quality assessment. The experimental results show that compared to alternative forced choice, the relaxed force choice with the not-sure option yielded a model with a better fit and a higher correlation between the proportions of correct responses and the physical scale. Moreover, the cognitive load assessment indicates that the \textit{Mental Demand} was significantly lower. In conclusion, the results of our study  support the use of the relaxed forced choice response format over the standard two-alternative forced choice. 

The results also show that the fitted distributions differed significantly between the two test conditions. This gives rise to a fundamental open research question, namely, to determine whether the improved performance of the RFC format also leads to a more accurate estimate for the collective JND in visual quality assessment studies. This is beyond the scope of our contribution and will be considered in future research. 

%We will see from the data of our experiment that this null hypothesis should be rejected. Thus, the question remains, which estimates, derived with or without the not-sure option, are more representative of the population means. However, while this problem is fundamental, it is outside the scope of this paper and will be addressed in another study.

\bibliographystyle{IEEEbib}
\bibliography{notsure}

\begin{thebibliography}{10}

\bibitem{testolina2022}
Michela Testolina,
\newblock ``Review of the state of the art on subjective image quality
  assessment,''
\newblock ISO/IEC, 2022,
\newblock Project ISO/IEC 29170 (JPEG AIC), document ISO/IEC JTC 1/SC 29/WG1
  N100163.

\bibitem{mantiuk2012comparison}
Rafa{\l}~K Mantiuk, Anna Tomaszewska, and Rados{\l}aw Mantiuk,
\newblock ``Comparison of four subjective methods for image quality
  assessment,''
\newblock in {\em Computer Graphics Forum}. Wiley Online Library, 2012,
  vol.~31, pp. 2478--2491.

\bibitem{fechner1889elemente}
Gustav~Theodor Fechner,
\newblock {\em Elemente der Psychophysik, zweite unveränderte Auflage},
  vol.~1,
\newblock Breitkopf und Hartel, 1889.

\bibitem{ISOIEC2015}
{ISO/IEC 29170-2},
\newblock ``{Information technology -- Advanced image coding and evaluation --
  Part 2: Evaluation procedure for visually lossless coding},'' 2015.

\bibitem{mcnally2017jpeg}
David McNally, Tim Bruylants, Alexandre Will{\`e}me, Touradj Ebrahimi, Peter
  Schelkens, and Benoit Macq,
\newblock ``{JPEG XS} call for proposals subjective evaluations,''
\newblock in {\em Applications of Digital Image Processing XL}. Intern.\ Soc.\
  Optics and Photonics, 2017, vol. 10396, pp. 103960P1--11.

\bibitem{watson2000proposal}
Andrew~B Watson,
\newblock ``Proposal: Measurement of a {JND} scale for video quality,''
\newblock {\em IEEE G-2.1. 6 Subcommittee on Video Compression Measurements},
  2000.

\bibitem{charrier2007maximum}
Christophe Charrier, Laurence~T Maloney, Hocine Cherifi, and Kenneth Knoblauch,
\newblock ``Maximum likelihood difference scaling of image quality in
  compression-degraded images,''
\newblock {\em JOSA A}, vol. 24, no. 11, pp. 3418--3426, 2007.

\bibitem{miao2008quantitative}
Jun Miao, Donglai Huo, and David~L Wilson,
\newblock ``Quantitative image quality evaluation of {MR} images using
  perceptual difference models,''
\newblock {\em Medical Physics}, vol. 35, no. 6, pp. 2541--2553, 2008.

\bibitem{menkovski2011value}
Vlado Menkovski, Georgios Exarchakos, and Antonio Liotta,
\newblock ``The value of relative quality in video delivery,''
\newblock {\em Journal of Mobile Multimedia}, vol. 7, no. 3, pp. 151--162,
  2011.

\bibitem{ponomarenko2013new}
Nikolay Ponomarenko, Oleg Ieremeiev, Vladimir Lukin, Lina Jin, Karen
  Egiazarian, Jaakko Astola, Benoit Vozel, et~al.,
\newblock ``A new color image database {TID2013}: Innovations and results,''
\newblock in {\em Advanced Concepts for Intelligent Vision Systems, ACIVS
  2013}, 2013, pp. 402--413.

\bibitem{kumcu2016performance}
Asli Kumcu, Klaas Bombeke, Ljiljana Plati{\v{s}}a, Ljubomir Jovanov, Jan
  Van~Looy, and Wilfried Philips,
\newblock ``Performance of four subjective video quality assessment protocols
  and impact of different rating preprocessing and analysis methods,''
\newblock {\em IEEE Journal of Selected Topics in Signal Processing}, vol. 11,
  no. 1, pp. 48--63, 2016.

\bibitem{nuutinen2016new}
Mikko Nuutinen, Toni Virtanen, Tuomas Leisti, Terhi Mustonen, Jenni Radun, and
  Jukka H{\"a}kkinen,
\newblock ``A new method for evaluating the subjective image quality of
  photographs: dynamic reference,''
\newblock {\em Multimedia Tools and Applications}, vol. 75, pp. 2367--2391,
  2016.

\bibitem{zhang2018unreasonable}
Richard Zhang, Phillip Isola, Alexei~A Efros, Eli Shechtman, and Oliver Wang,
\newblock ``The unreasonable effectiveness of deep features as a perceptual
  metric,''
\newblock in {\em Proceedings of the IEEE Conference on Computer Vision and
  Pattern Recognition}, 2018, pp. 586--595.

\bibitem{sun2017mdid}
Wen Sun, Fei Zhou, and Qingmin Liao,
\newblock ``{MDID: A} multiply distorted image database for image quality
  assessment,''
\newblock {\em Pattern Recognition}, vol. 61, pp. 153--168, 2017.

\bibitem{men2021subjective}
Hui Men, Hanhe Lin, Mohsen Jenadeleh, and Dietmar Saupe,
\newblock ``Subjective image quality assessment with boosted triplet
  comparisons,''
\newblock {\em IEEE Access}, vol. 9, pp. 138939--138975, 2021.

\bibitem{punch2001paired}
Jerry~L Punch, Brad Rakerd, and Amyn~M Amlani,
\newblock ``Paired-comparison hearing aid preferences: Evaluation of an
  unforced-choice paradigm,''
\newblock {\em Journal of the American Academy of Audiology}, vol. 12, no. 04,
  pp. 190--201, 2001.

\bibitem{kaernbach2001adaptive}
Christian Kaernbach,
\newblock ``Adaptive threshold estimation with unforced-choice tasks,''
\newblock {\em Perception \& Psychophysics}, vol. 63, no. 8, pp. 1377--1388,
  2001.

\bibitem{kaernbach1991simple}
Christian Kaernbach,
\newblock ``Simple adaptive testing with the weighted up-down method,''
\newblock {\em Perception \& Psychophysics}, vol. 49, no. 3, pp. 227--229,
  1991.

\bibitem{garcia2017indecision}
Miguel~A Garc{\'\i}a-P{\'e}rez and Roc{\'\i}o Alcal{\'a}-Quintana,
\newblock ``The indecision model of psychophysical performance in
  dual-presentation tasks: Parameter estimation and comparative analysis of
  response formats,''
\newblock {\em Frontiers in Psychology}, vol. 8, pp. 1142, 2017.

\bibitem{garcia2019Thedos}
Miguel~{\'A}ngel Garc{\'\i}a-P{\'e}rez and Roc{\'\i}o Alcal{\'a}-Quintana,
\newblock ``The do’s and don’ts of psychophysical methods for
  interpretability of psychometric functions and their descriptors,''
\newblock {\em The Spanish Journal of Psychology}, vol. 22, 2019.

\bibitem{horton2010dot}
John~J Horton,
\newblock ``The dot-guessing game: A ‘fruit fly’ for human computation
  research,''
\newblock {\em Available at SSRN 1600372}, pp. 1--6, 2010.

\bibitem{honda2022round}
Hidehito Honda, Rina Kagawa, and Masaru Shirasuna,
\newblock ``On the round number bias and wisdom of crowds in different response
  formats for numerical estimation,''
\newblock {\em Scientific Reports}, vol. 12, no. 1, pp. 1--18, 2022.

\bibitem{ugander2015wisdom}
Johan Ugander, Ryan Drapeau, and Carlos Guestrin,
\newblock ``The wisdom of multiple guesses,''
\newblock in {\em Proceedings of the Sixteenth ACM Conference on Economics and
  Computation}, 2015, pp. 643--660.

\bibitem{kemmer2020enhancing}
Ryan Kemmer, Yeawon Yoo, Adolfo Escobedo, and Ross Maciejewski,
\newblock ``Enhancing collective estimates by aggregating cardinal and ordinal
  inputs,''
\newblock in {\em Proceedings of the AAAI Conference on Human Computation and
  Crowdsourcing}, 2020, vol.~8, pp. 73--82.

\bibitem{pfeiffer2012adaptive}
Thomas Pfeiffer, Xi~Alice Gao, Yiling Chen, Andrew Mao, and David~G Rand,
\newblock ``Adaptive polling for information aggregation,''
\newblock in {\em Twenty-Sixth AAAI Conference on Artificial Intelligence},
  2012, pp. 122--128.

\bibitem{hart1988development}
Sandra~G Hart and Lowell~E Staveland,
\newblock ``Development of nasa-tlx (task load index): Results of empirical and
  theoretical research,''
\newblock in {\em Advances in Psychology}, vol.~52, pp. 139--183. Elsevier,
  1988.

\bibitem{10.1145/3582272}
Thomas Kosch, Jakob Karolus, Johannes Zagermann, Harald Reiterer, Albrecht
  Schmidt, and Pawe\l{}~W. Wo\'{z}niak,
\newblock ``A survey on measuring cognitive workload in human-computer
  interaction,''
\newblock {\em ACM Computing Surveys}, pp. 1--37, 2023.

\bibitem{prins2016Psychophysik}
Frederick~A.A. Kingdom and Nicolaas Prins,
\newblock {\em {Psychophysik: Eine praktische Einführung}},
\newblock Academic Press, 2016.

\bibitem{lakens2013calculating}
Dani{\"e}l Lakens,
\newblock ``Calculating and reporting effect sizes to facilitate cumulative
  science: a practical primer for t-tests and {ANOVAs},''
\newblock {\em Frontiers in Psychology}, vol. 4, pp. 863, 2013.

\bibitem{wichmann2001psychometric}
Felix~A Wichmann and N~Jeremy Hill,
\newblock ``The psychometric function: {I. Fitting, sampling, and goodness of
  fit},''
\newblock {\em Perception \& Psychophysics}, vol. 63, no. 8, pp. 1293--1313,
  2001.

\bibitem{garcia2018nonparametric}
Miguel~A Garc{\'\i}a-P{\'e}rez and Vicente N{\'u}{\~n}ez-Ant{\'o}n,
\newblock ``Nonparametric tests for equality of psychometric functions,''
\newblock {\em Behavior Research Methods}, vol. 50, pp. 2226--2255, 2018.

\end{thebibliography}

\end{document}